\documentclass{article}

\usepackage[main, final]{neurips_2026}
\makeatletter
\renewcommand{\@trackname}{}
\makeatother

\usepackage[utf8]{inputenc} 
\usepackage[T1]{fontenc}    
\usepackage{hyperref}       
\usepackage{url}            
\usepackage{booktabs}       
\usepackage{amsfonts}       
\usepackage{nicefrac}       
\usepackage{microtype}      
\usepackage{xcolor}         

\usepackage{amsmath}
\usepackage{bm}
\usepackage{derivative}
\usepackage{graphicx}
\usepackage{subcaption}
\usepackage{tikz}
\usetikzlibrary{positioning,fit}
\usepackage{pdflscape}

\newcommand{\real}{\mathbb{R}}

\newcommand{\expectation}{\mathbb{E}}
\newcommand{\calX}{\mathcal{X}}
\newcommand{\calY}{\mathcal{Y}}

\newcommand{\pdatadef}{p_{\text{data}}}
\newcommand{\pnoise}{p_{\text{noise}}}

\title{Joint Flow Matching: Consistent Generative and Discriminative Inference from a Shared Flow}

\author{%
  Hayden ~McAlister\\
  School of Computing\\
  University of Otago\\
  Dunedin, New Zealand \\
  \texttt{hayden.mcalister@otago.ac.nz} \\
  \And
  Lech ~Szymanski\\
  School of Computing\\
  University of Otago\\
  Dunedin, New Zealand \\
  \texttt{lech.szymanski@otago.ac.nz} \\
}

\begin{document}

\maketitle

\begin{abstract}
We study Joint Flow Matching (JFM), a flow-based framework for learning a joint distribution over paired variables such as images and class labels. JFM transports the variables in opposite directions: an image is transported from noise to data while its label is transported from data to noise. This provides a single probabilistic model in which generation and classification are represented as complementary conditional inferences. We formulate the joint distribution induced by this construction and characterise the relationship between its forward and reverse conditionals. We investigate the resulting model for classification and conditional generation, including confidence calibration and consistency between generated samples and classifier predictions. Experiments on conditional datasets demonstrate competitive classification accuracy, well-calibrated confidence without post-hoc calibration, and classifier-consistent generation. We further explore the shared joint distribution as a basis for interpreting the relationship between discriminative and generative behaviour.
\end{abstract}

\section{Introduction}

The goal of any generative model is to learn a mapping from a tractable noise distribution to a complex data distribution. Continuous normalising flows (CNFs) \citep{ChenEtAl2018} accomplish this by learning a time-indexed velocity field that transports probability mass from noise at $t=0$ to data at $t=1$. Flow Matching \citep{LipmanEtAl2023} has emerged as a prominent training paradigm for CNFs, enabling scalable training by forgoing the need for costly numerical integration at each epoch. Most generative models (including CNFs) treat generation as a one-directional process: noise samples are pushed forward to approximate a data distribution. When the data consists of paired variables $(X, Y)$ (e.g., images and class labels), these models typically capture $p(X|Y)$. Sampling the conditional distribution $p(Y | X)$ requires either a separate discriminative model or costly approximate inference. There is no guarantee of consistency between the two directions.

We present \emph{Joint Flow Matching} (JFM), a mechanism that addresses this asymmetry by assigning noise and data distributions to opposite temporal endpoints: $p_0(X, Y) = \pnoise(X)\pdatadef(Y)$ and $p_1(X, Y) = \pdatadef(X)\pnoise(Y)$. A single reversible flow between these endpoints supports two conditional directions:
\begin{itemize}
    \item \textbf{Generation} ($t: 0 \to 1$): given a label $y_0 \sim \pdatadef(Y)$, integrate forward to obtain image $x_1 \sim \pdatadef(X)$ conditioned on the label.
    \item \textbf{Classification} ($t: 1 \to 0$): given an image $x_1 \sim \pdatadef(X)$, integrate backward to obtain label $y_0$ conditioned on the image.
    \item \textbf{Consistency}: both conditional directions are induced by the same joint distribution and are therefore mutually consistent.
\end{itemize}

\begin{figure}[t!]
    \centering
    \begin{minipage}{0.48\textwidth}
        \begin{subfigure}[t]{\textwidth}
            \centering
            \includegraphics[width=\textwidth]{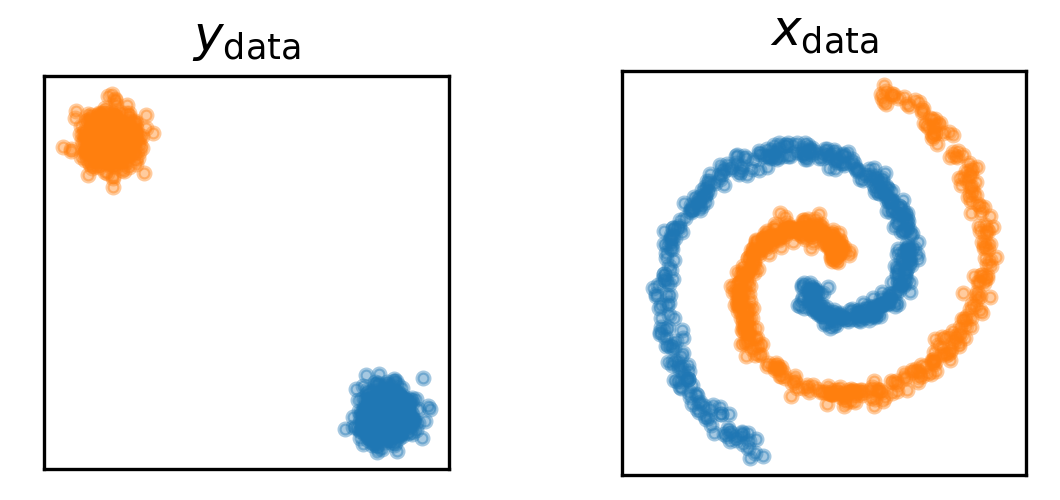}
            \caption{The label and data distributions.}
            \label{subfigure: spiral data}
        \end{subfigure}
        \begin{subfigure}[t]{\textwidth}
            \centering
            \includegraphics[width=\textwidth]{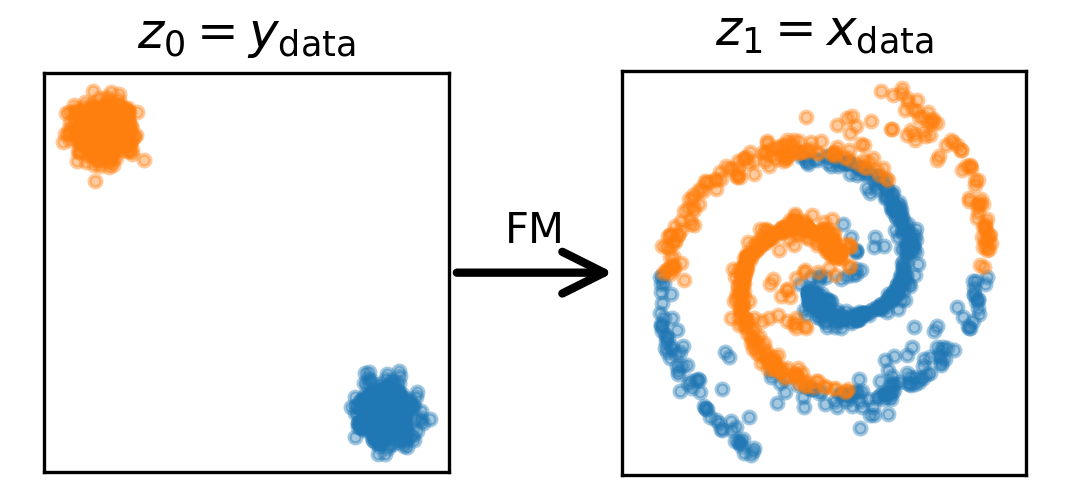}
            \caption{Standard flow matching generation.}
            \label{subfigure: spiral fm}
        \end{subfigure}
    \end{minipage}%
    ~
    \begin{minipage}{0.55\textwidth}
        \begin{subfigure}[t]{\textwidth}
            \centering
            \includegraphics[width=\textwidth]{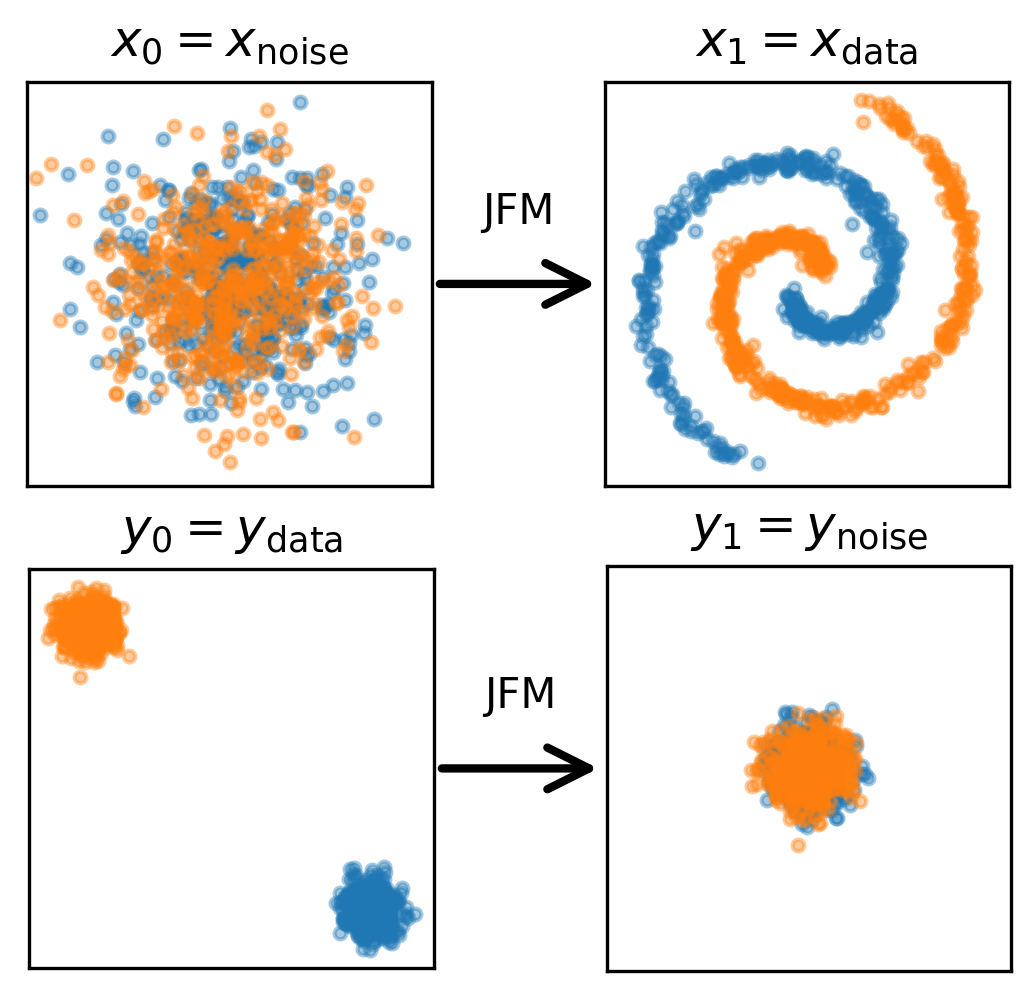}
            \caption{Joint flow matching generation.}
            \label{subfigure: spiral jfm}
        \end{subfigure}
    \end{minipage}

    \caption{Forward flow in FM and JFM trained to model $p(x|y)$ and $p(y|x)$ for the dataset shown in (\subref{subfigure: spiral data}). Colour indicates the class conditioning, corresponding to the distinct arms of the spiral. FM can in principle map arbitrary source and target distributions but flowing $y$ directly to $x$ to enable bidirectional conditioning performs poorly for the spiral under the linear interpolation path. (\subref{subfigure: spiral fm}). JFM has no trouble recovering $p(x|y)$ under the same transport (\subref{subfigure: spiral jfm}). Reverse flow for the two models is shown in Appendix \ref{appendix: reverse spiral}.}
    \label{figure: spiral fm vs jfm}
\end{figure}

This reversible conditional flow matching mechanism was also proposed by \citet{Caetano.etal2026} under the name \textit{Symmetrical Flow Matching} (SFM). While SFM explores the generative and classification capabilities of the unified model, our work focuses on the joint distribution induced by the construction.  We establish theoretical guarantees for the consistency of its generative and discriminative conditionals. We then demonstrate the conditional-generation behaviour of JFM compared to standard FM using the spiral dataset (Figure \ref{figure: spiral fm vs jfm}). Finally, we show that the consistency between the generative and discriminative conditionals enables functionality beyond generation and classification on real image datasets, including calibrated predictions and new avenues for model interpretability.

This work is complementary to SFM, which established that a reverse-coupled flow can support unified generation and classification \citep{Caetano.etal2026}. The motivation for JFM is not that a jointly trained model should outperform separately trained classifiers or generators, but that a shared joint model provides a relationship between generation and classification that is absent when the two are trained independently. By deriving both directions of inference from the same model, JFM allows generated samples to be evaluated by the corresponding classifier and class-conditioned transformations to be examined for their effect on the model's predictions. These capabilities provide a model-internal means of examining the learned relationships without requiring prior specification of features or failure modes.

\paragraph{Contributions}
\begin{enumerate}
\item We formulate Joint Flow Matching as a joint probabilistic model over paired variables $(X,Y)$, providing a formal probabilistic characterisation of the shared reversible flow underlying generative and discriminative inference.
\item We show theoretically that the flow construction yields consistency between the conditional distributions $p_\text{JFM}(X\mid Y)$ and $p_\text{JFM}(Y\mid X)$, without requiring additional consistency constraints or regularisation.
\item We empirically investigate the implications of this consistency for classification, conditional generation, and their relationship.
\end{enumerate}

\begin{figure}[t!]
    \centering
    \includegraphics[width=0.95\textwidth]{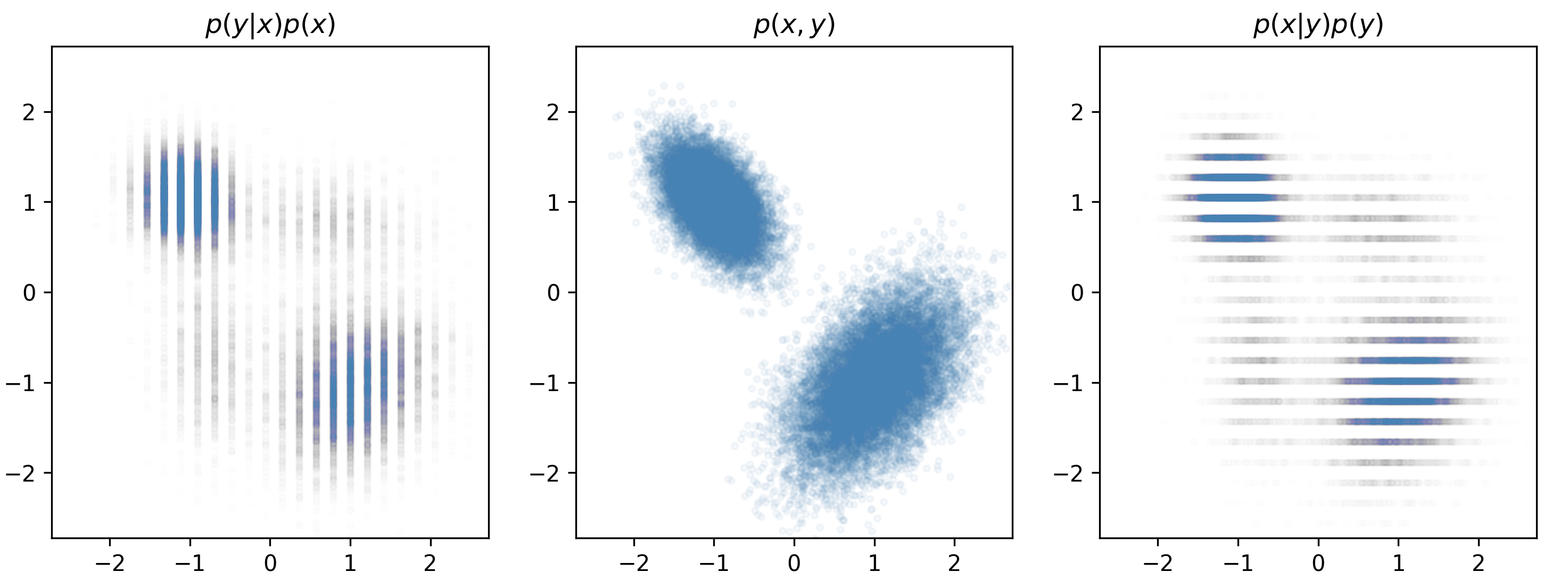}
    \caption{Illustration of the consistency of JFM. A flow is trained on the marginal distributions along the X and Y axes. The central plot shows the true joint distribution of the two variables. The left and right plots show the generated joint when conditioning on the X or Y variables respectively. Note the shadows in the upper right and lower left quadrants of the generated distributions; although not present in the true distribution they are consistent between the generated marginals -- generated joints match. The bands in the joints are for visualisation purposes only to make it clear which axis has been sampled over.}
    \label{figure: sliced joint example}
\end{figure}

\section{Theory}
\label{section: theory}

\subsection{Setup and Endpoint Distributions}

Consider a joint space $\calX \times \calY = \real^{|\calX|} \times \real^{|\calY|}$ with a temporal parameter $t \in [0, 1]$. Let $\pdatadef(X)$ and $\pdatadef(Y)$ denote the marginal data distributions, with tractable noise distributions $\pnoise(X)$ and $\pnoise(Y)$ defined over the same spaces. We define the endpoint distributions as:
\begin{equation}
    p_0(x, y) = \pnoise(x)\,\pdatadef(y), \qquad p_1(x, y) = \pdatadef(x)\,\pnoise(y).
    \label{equation: endpoints}
\end{equation}
Both are valid joint distributions: at each endpoint, the two factors are independent by construction, since the noise distribution is chosen independently of the data distribution. Unlike standard CNFs, in which noise and data are assigned uniformly across variables at $t=0$ and $t=1$ respectively, here they are assigned in opposite roles for both $X$ and $Y$ \citep{Caetano.etal2026}.

We learn a continuous normalising flow $\Phi_t : \calX \times \calY \to \calX \times \calY$ defined by a velocity field $v$:
\begin{equation}
    \odv{}{t}\Phi_t(x, y) = v(t, \Phi_t(x, y)), \qquad \Phi_0(x, y) = (x, y).
    \label{equation: flow}
\end{equation}
Under regularity conditions on $v$, $\Phi_t$ is a diffeomorphism for each $t$, and in particular $\Phi_1$ is a bijection. The flow is trained so that $(\Phi_1)_* p_0 = p_1$, i.e.\ $\Phi_1$ pushes $p_0$ forward to $p_1$. By the change of variables formula,
\begin{equation}
    p_1(x_1, y_1) = p_0\!\left(\Phi_1^{-1}(x_1, y_1)\right) \cdot \left|\det J_{\Phi_1^{-1}}(x_1, y_1)\right|,
    \label{equation: pushforward}
\end{equation}
where $J_{\Phi_1^{-1}}$ denotes the Jacobian of $\Phi_1^{-1}$. Concretely, the transported sample $(x_1, y_1) = \Phi_1(x_0, y_0)$ is obtained by integrating the ODE from Equation~\ref{equation: flow}:
\begin{equation}
    (x_1, y_1) = (x_0, y_0) + \int_0^1 v(t, x_t, y_t)\, dt.
    \label{equation: sample-pushforward}
\end{equation}

\subsection{The Joint Distribution}

Standard CNFs couple variables at a shared data endpoint ($t=1$), making the joint $p_1(x, y) = \pdatadef(x, y)$ explicit. In JFM the endpoint distributions are factored by design, so the learned coupling between $X$ and $Y$ is not directly visible at either end. It is instead encoded in the joint $p_\text{JFM}(X, Y)$, defined as the joint distribution of the pair $(x_1, y_0)$: the $\calX$-output of the forward flow paired with the $\calY$-input, or equivalently the $\calY$-output of the reverse flow paired with the $\calX$-input.

Formally, we define the map $\Psi : \calX \times \calY \to \calX \times \calY$ by
\begin{equation}
    \Psi(x_0, y_0) = \bigl(\pi_x(\Phi_1(x_0, y_0)),\; y_0\bigr),
    \label{equation: psi}
\end{equation}
where $\pi_x$ denotes the projection onto $\calX$ by discarding the $\calY$ variable. The joint distribution is the pushforward $p_\text{JFM} := \Psi_* p_0$. We assume that for almost every $y_0 \in \calY$ the map $\pi_x \circ \Phi_1(\cdot, y_0) : \calX \to \calX$ is a diffeomorphism. Under this assumption, $\Psi$ is itself a diffeomorphism, so the change of variables formula applies. Since $\Psi$ transforms $x_0 \mapsto x_1$ while holding $y_0$ fixed, its Jacobian is block-triangular:
\begin{equation*}
    J_\Psi(x_0, y_0) = \begin{pmatrix} \partial x_1 / \partial x_0 & \partial x_1 / \partial y_0 \\ 0 & I \end{pmatrix},
\end{equation*}
so $|\det J_\Psi| = |\det \partial x_1 / \partial x_0|$. Applying the change of variables formula:
\begin{equation}
    p_\text{JFM}(x_1, y_0) = p_0(x_0, y_0) \left| \det \frac{\partial x_0}{\partial x_1} \bigg|_{y_0} \right| = \pnoise(x_0) \pdatadef(y_0) \left| \det \frac{\partial x_0}{\partial x_1} \big|_{y_0} \right|,
    \label{equation: joint distribution}
\end{equation}
where $x_0 = \bigl( \pi_x \circ \Phi_1(\cdot, y_0) \bigr)^{-1}(x_1)$. The Jacobian factor is replaced by a partial Jacobian for fixed $y_0$. The same derivation can be made using the reverse flow starting with fixed $x_1$ and finding $y_1 \mapsto y_0$.

While $\Phi_1$ is a diffeomorphism on the full product space $\calX \times \calY$ by the regularity of $v$, this does not in general imply that the slice maps $\pi_x \circ \Phi_1(\cdot, y_0) : \calX \to \calX$ are diffeomorphisms. The slice diffeomorphism assumption is not derivable from the ODE regularity conditions alone, but it is consistent with the inductive bias of neural velocity fields: a violation would require two distinct noise inputs, at the same label $y_0$, to produce the same data output $x_1$ under the learned flow. However, we almost never (in the statistical sense) observe this in practical neural velocity fields.

\subsection{Consistency of Generative and Discriminative Conditionals}

The central claim of JFM is that forward and backward integration through the same trained flow yield samples from the two conditionals of a single joint distribution $p_\text{JFM}$, guaranteeing consistency between generation and classification.

\begin{itemize}
    \item \textbf{Forward (generation):} draw $x_0 \sim \pnoise(X)$ and $y_0 \sim \pdatadef(Y)$, integrate forward to obtain $(x_1, y_1) = \Phi_1(x_0, y_0)$. The output $x_1$ is a sample from $p_\text{JFM}(x_1 | y_0)$.
    \item \textbf{Backward (classification):} given $x_1 \sim \pdatadef(X)$, draw $y_1 \sim \pnoise(Y)$ and integrate backward to obtain $(x_0, y_0) = \Phi_1^{-1}(x_1, y_1)$. The output $y_0$ is a sample from $p_\text{JFM}(y_0 | x_1)$.
\end{itemize}

Both claims follow directly from the definition of $p_\text{JFM}$ as $\Psi_* p_0$. In the forward direction, $x_1 = \pi_x(\Phi_1(x_0, y_0))$, so for fixed $y_0$ the distribution of $x_1$ is exactly $p_\text{JFM}(x_1 | y_0)$. In the backward direction, fixing \(x_1\) and marginalising over \(y_1 \sim \pnoise(Y)\) via \(\Phi_1^{-1}\) recovers the same conditional \(p_\text{JFM}(y_0 \mid x_1)\), since \(\Phi_1\) is bijective and maps the prescribed source distribution to the prescribed target distribution. Bayes' theorem then gives the consistency relation:

\begin{equation}
    p_\text{JFM}(x_1 | y_0)\,\pdatadef(y_0) = p_\text{JFM}(y_0 | x_1)\,\pdatadef(x_1).
    \label{equation: consistency}
\end{equation}

This relationship holds for any flow $\Phi_1$ satisfying the endpoint conditions in Equation~\ref{equation: endpoints}. In standard CNFs both variables are transported from noise to data simultaneously, and there is no analogous backward pass that isolates $p(Y | X)$: fixing $x_1$ and integrating backward recovers $x_0 \sim \pnoise(X)$ but gives no information about $y$. There is no way to decompose the combined data distribution, which JFM solves by isolating one variable at either end of the time domain. 

While the theoretical consistency result is formally established only for continuous $Y$, our experiments suggest that the construction can be applied effectively to categorical variables in practice. A formal treatment of discrete $Y$, including continuous relaxations or simplex-based formulations, is left for future work. 

\subsection{Training Objective}\label{sub_training}

Following standard Flow Matching \citep{LipmanEtAl2023}, we express the velocity field as a parametric model and train it by minimising a conditional flow matching loss. For each training pair $(x_1, y_0) \sim \pdatadef(X, Y)$ we sample noise variables $x_0 \sim \pnoise(X), y_1 \sim \pnoise(Y)$ and time $t \sim \mathcal{U}(0, 1)$.

We then form the linearly interpolated intermediate states and their corresponding conditional target velocities:
\vspace{-0.4cm}
\begin{align*}
    x_t &= (1 - t) x_0 + t x_1, & u_x &= x_1 - x_0, \\
    y_t &= (1 - t) y_0 + t y_1, & u_y &= y_1 - y_0.
\end{align*}
Note that $u_x = x_1 - x_0$ is a positive-direction target velocity (noise to data), while $u_y = y_1 - y_0$ is a negative-direction target (data to noise). This is due to the reversed roles of the Y variable inherent to JFM. Using the conditional flow matching trick proposed by \citet{LipmanEtAl2023} and the optimal transport probability path to interpolate velocities across time we construct the training loss:
\begin{equation}
    \mathcal{L}(\theta) = \expectation_{\begin{subarray}{l}
        t \sim \mathcal{U}[0, 1] \\
        (x_1, y_0) \sim \pdatadef(X, Y) \\
        x_0 \sim \pnoise(X), y_1 \sim \pnoise(Y)
    \end{subarray}}
    \|v_x^\theta(x_t, y_t, t) - u_x(x_t)\|^2 + \|v_y^\theta(x_t, y_t, t) - u_y(y_t)\|^2.
    \label{equation: JFM loss}
\end{equation}

\section{Related Work}

There are many families of generative models. Generative adversarial networks \citep{GoodfellowEtAl2014} train a generator and discriminator jointly via a minimax objective. While capable of high-quality synthesis, training a GAN is often unstable \citep{SrivastavaEtAl2017, Thanh-TungTran2020}, the generator and discriminator are architecturally separate, and there is no mechanism for consistent conditional inference. Variational autoencoders \citep{KingmaEtAl2014} introduce an encoder-decoder pair trained by modelling the latent-space distribution, but are limited in both sample quality and density estimation accuracy. Diffusion models \citep{Sohl-DicksteinEtAl2015, HoEtAl2020, SongEtAl2021, YangEtAl2023} achieve state-of-the-art sample quality by learning to reverse a fixed forward-noising process and have been extensively studied for their generalisation properties \citep{KadkhodaieEtAl2024, NiedobaEtAl2025, Vastola2025}.  

Discrete normalising flows \citep{TabakTurner2013, RezendeMohamed2016, DinhEtAl2017, KingmaEtAl2017} construct a bijection from a tractable base distribution to the data distribution using a physically inspired velocity field. The bijective nature of normalising flows gives rise to exact likelihood computation, although discrete normalising flows require expensive computation for training and inference. Continuous normalising flows (CNFs) \citep{ChenEtAl2018, GrathwohlEtAl2018} generalise discrete normalising flows to a continuous-time setting, parameterising the dynamics of a probability density over time and using a neural ODE to model them. Maximum-likelihood training of CNFs requires expensive simulation of the ODE during each gradient step; Flow Matching \citep{LipmanEtAl2023} resolves this by forming a tractable conditional velocity field with equivalent gradient once marginalised.



\subsection{Generation-Classification Duality}

Several lines of work have sought to unify generation and classification in a single model. Energy-based models \citep{DuMordatch2019, SongKingma2021} define a scalar energy $E(x, y)$ from which both $p(x)$ and $p(y | x)$ can be derived. However, inference requires expensive MCMC methods and uses an intractable partition function, making exact likelihoods impossible. Joint energy-based models such as JEM \citep{GrathwohlEtAl2020} reinterpret a discriminative classifier as an energy-based generative model, but generation quality suffers compared to dedicated generative models and MCMC remains a bottleneck \citep[although this was addressed in][]{GrathwohlEtAl2020a}. Classifier-free guidance \citep{HoSalimans2022} steers diffusion samples toward a desired class at inference time but fails to produce a consistent joint model. 

\citet{LiEtAl2023}, building on the work of \citet{ClarkJaini2023}, construct a classifier from a diffusion generator by comparing the quality of multiple restorations of an image corrupted with noise, each time conditioning on a different candidate from a set of classification labels. Though re-purposing of the generator is a useful trick, it does not give actual consistency between generation and classification. CycleGAN \citet{ZhuEtAL2017} employs a similar training method to our FM setup in the spiral example in Figure \ref{figure: spiral fm vs jfm}, learning a direct mapping from one conditional to another. Their work can be used to map from images conditioned on one label to those conditioned on another. JFM can achieve this, as illustrated in Figure \ref{figure: cifar10 class swap experiment}, in which we switch the conditional $y_0$ while keeping the noise $x_0$ derived from an image, producing new images that swap the object retaining independent context. The consistency property of JFM means we use this visualisation to show what is correlated and what is not with the class membership of the model.

Classical saliency methods (\citealp[e.g.][]{SimonyanEtAl2014, ZeilerFergus2014}, \citealp[see][for a metastudy]{LiEtAl2021}) compute input-gradient maps or feature activation visualisations for a fixed classifier, but these are separate from any generative model. GAN-based approaches \citep{GoetschalckxEtAl2019} explore the latent geometry of the generator model but require a separately trained discriminator to assign meaning to generated images. Integrated gradients \citep{SundararajanEtAl2017} and Shapley-value methods \citep{LundbergLee2017} give semantic attributions to inputs for classifiers but are decoupled from any generative model. The joint distribution in JFM arises naturally from the coupling learned by the flow itself. The attributions are drawn from the same model that performs both generation and classification.

\citet{Caetano.etal2026} proposed Symmetrical Flow Matching (SFM), which also uses opposite endpoint assignments for image and label variables to enable both generative and discriminative inference within a single flow model. SFM develops this construction as a unified framework for image generation, semantic segmentation, and classification, with an emphasis on task performance and efficient bidirectional inference. In contrast, JFM focuses on the probabilistic structure induced by the shared flow and formally characterises the consistency between the corresponding generative and discriminative conditionals. JFM also uses a simpler label-flow construction in which the label velocity is predicted from the bottleneck representation rather than propagated through the full image-processing network. With this formulation, JFM provides stable reverse classification across inference step counts without requiring label perturbation.

\section{Methodology}
\label{section: methodology}

The JFM framework can be implemented using any parametric velocity model $v(t, x_t, y_t)$ and trained using the flow matching objective set out in Equation \ref{equation: JFM loss}. The exact form of the velocity model will almost certainly be determined by the application. For simple toy examples (see Section \ref{subsection: spiral results}) we have employed a small multilayer-perceptron with X and Y concatenated for input and output. All training runs used AdamW optimiser with default weight decay of 0.01 and learning rate decayed with cosine annealing.

\subsection{Image Generation-Classification}

For image generation-classification we used an ADM-style U-Net backbone \citep{DhariwalNichol2021}, following prior diffusion and flow-matching work, with the key addition of a classification head attached at the bottleneck for label velocity output. A detailed schematic of the architecture and its dimensions is provided in Appendix~\ref{appendix: unet}.

The U-Net takes as input the image $x_t \in \real^{C \times H \times W}$, label $y_t \in \real^K$, and time $t \in [0, 1]$, and produces output velocities $(v_x(t), v_y(t)) = v(x_t, y_t, t)$. The image is fed end-to-end with standard encoder and decoder made up of attention and residual blocks \citep{HeEtAl2015}. Time $t$ is sinusoidally embedded \citep{VaswaniEtAl2017} and injected into the model along with the embedding of the label $y_t$ via scale-shift normalisation \citep{DhariwalNichol2021}. Image velocity $v_x(t)$ is taken at the output of the decoder network, which naturally has the same dimension as the input image. 

We extract the label velocity $v_y(t)$ from the bottleneck of the U-Net after passing it through an MLP classifier head. The bottleneck encodes the most abstract, class-discriminative features of the image, making it the natural place to extract information about the label. This mirrors the intuition that classification relies on global, low-frequency image structure, which is most concentrated at the coarsest spatial scale. 

\subsection{Training}

We trained U-Net to model the objective described in Section \ref{sub_training} using the loss defined in Equation \ref{equation: JFM loss} with  earning rate of $1\times 10^{-4}$. Images $x_1$ were standardised to zero mean and unit variance, the labels $y_0$ one-hot encoded. Noise samples $x_0$ and $y_1$ were sampled from the multivariate Gaussian distribution, $x_0\sim \mathcal{N}(0,\sigma^2_x I)$ and $y_1\sim \mathcal{N}(0,\sigma^2_y I)$, with $\sigma_x=1.0$ and $\sigma_y=0.1$.

The best velocity model weights were selected based on validation classification accuracy, evaluated after each epoch using a single 5-step reverse-flow pass over validation data with fixed noise initialisation of $y_1=\mathbf{0}$. 

\subsection{Inference}

\paragraph{Generation ($t: 0 \to 1$)} To generate an image conditioned on label $y_0$, we sample image noise $x_0 \sim \pnoise(X)$ and numerically integrate the ODE from Equation \ref{equation: flow} forward from $t = 0$ to $t = 1$ using the Euler method:
\begin{equation*}
    x_{t+\Delta t} = x_t + \Delta t\cdot v_x^\theta(x_t, y_t, t), \qquad
    y_{t+\Delta t} = y_t + \Delta t\cdot v_y^\theta(x_t, y_t, t).
\end{equation*}
The output $x_1$ is the generated image; $y_1$ is discarded. Conditioning is implicit on the label $y_0$ which starts as a one-hot encoded class label.

\paragraph{Classification ($t: 1 \to 0$)} To classify an image $x_1$, we initialise $y_1 \sim \pnoise(Y)$ and integrate the ODE backward from $t = 1$ to $t = 0$ by the same process as above with negated velocity:
\begin{equation*}
    x_{t-\Delta t} = x_t - \Delta t\cdot v_x^\theta(x_t, y_t, t), \qquad
    y_{t-\Delta t} = y_t - \Delta t\cdot v_y^\theta(x_t, y_t, t).
\end{equation*}
The output $y_0$ is a real-valued score vector over the classes. The predicted class is taken to be $\hat{k} = \text{argmax}_k (y_0)_k$. We estimate classification confidence via Monte Carlo sampling, drawing multiple samples of $y_1$ with the same $x_1$, flowing it to $y_0$ and taking the most frequent predicted label; confidence is its empirical frequency.

\section{Results}
\label{section: results}

 \label{subsection: cifar10}

\begin{figure}[t!]
    \centering
    \includegraphics[width=0.85\textwidth]{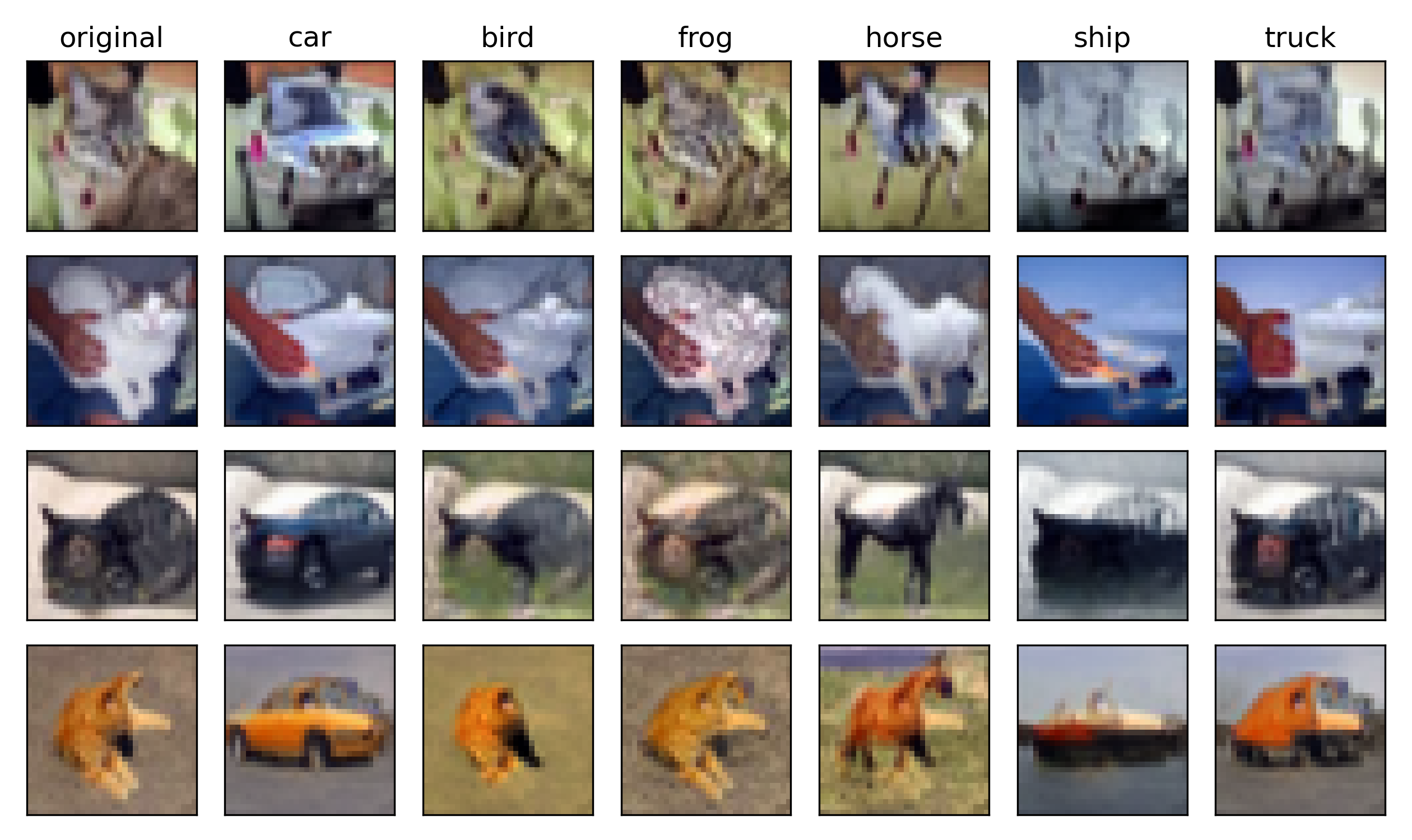}
    \caption{CIFAR-10 images JFM-generated from the noise $x_0$ recovered via reverse flow from the original image from the class ``cat" (left column) then forward flowed with different label conditioning in each following column.}
    \label{figure: cifar10 class swap experiment}
\end{figure}
\subsection{The Spiral}
\label{subsection: spiral results}

We generated this dataset from a bimodal distribution whose modes lie along two interleaving spiral curves. The labels correspond to the modes of the distribution one-hot encoded in $\real^2$. With some added noise, these form $\pdatadef(X)$ and $\pdatadef(Y)$ respectively. Both standard flow matching and joint flow matching were trained with an MLP velocity model of four hidden layers of $128$ SiLU units each. We trained over 200 epochs with the batch size of 1024 and the starting learning rate of $10^{-3}$. ODE integration uses the Euler method with $64$ function evaluations. The results are shown in Figures \ref{figure: spiral fm vs jfm} and \ref{figure: reverse spiral fm vs jfm}.

\paragraph{FM.} The flow is trained to transport labels $z_0 \sim \pdatadef(Y)$ to data $z_1 \sim \pdatadef(X)$. For this, FM necessitates that both variables are embedded in the shared space and the labels are infused with some noise (as FM's bijective flow would not be able to map two point labels to more points on the spiral). Each class occupies a mode of a multimodal Gaussian at $t=0$, and the flow must scatter these clusters onto the correct spiral arm at $t=1$. Though not typical, FM does allow this. However, the spiral arms are tightly intertwined: straight-line paths from a label cluster to its target arm inevitably pass through the opposite arm. This could be solved with a flow over appropriate manifold, but we assume no such prior knowledge, in which case straight-line paths are preferred by the optimal transport assumptions \citep{ChenLipman2024}. As shown in Figure \ref{subfigure: spiral fm}, the resulting samples populate both arms regardless of the conditioning label: $\pdatadef(X)$ is recovered, but $\pdatadef(X \mid Y)$ is essentially uninformative.

\paragraph{JFM.} Each data sample $x_1 \sim \pdatadef(X)$ is paired with noise $x_0\sim \mathcal{N}(0,I)$ and label sample $y_0 \sim \pdatadef(Y)$ with $y_1\sim \mathcal{N}(0,0.1\cdot I)$ for linear interpolation and trained jointly over $t\sim\mathcal{U}(0,1)$. Because the label and data variables occupy separate spaces, the label signal is never corrupted by proximity to the data manifold, and remains informative throughout integration even as it is transported toward noise. The flow (in the forward direction) simultaneously routes $x_0$ to the correct spiral arm and transports $y_0$ to $\pnoise(Y)$, decoupling class conditioning from intra-class variability. As shown in Figure \ref{subfigure: spiral jfm}, JFM achieves near-perfect conditioning fidelity: samples conditioned on two class flow to appropriate arms of the spiral. In Figure \ref{subfigure: reverse spiral jfm}, the backward pass recovers the correct class label from a data point, confirming that a single model simultaneously performs generation and classification.

\subsection{Image Data}

We demonstrate implications of JFM's consistency using the CIFAR-10 dataset \citep{KrizhevskyHinton2009}. We use the modified U-Net architecture described in Section \ref{section: methodology}. Our model was trained on an NVIDIA L40 GPU for 500 epochs with standard data augmentation. We perform classification via reverse ODE integration over $20$ steps. Confidence is defined as the empirical frequency of the modal prediction across 10 samples of label noise $y_1$, and calibration was evaluated using top-label Expected Calibration Error (ECE) with 10 equal-width bins. Resulting test accuracy was $0.907$ with ECE of $0.058$ (see Appendix~\ref{appendix: cifar10 results} for a sensitivity analysis). On the generative side, using 200-step ODE sampling against the CIFAR-10 training distribution, we obtain an FID-50k score of $33.55$. This generative performance is below that of specialised CIFAR-10 generative models, reflecting the trade-off involved in using a shared model for both generation and classification.

Figure \ref{figure: cifar10 class swap experiment} shows conditional reconstruction akin to CycleGAN \citet{ZhuEtAL2017}, which takes an explainability aspect in JFM. We take a test image from CIFAR-10 dataset, and reverse flow with $y_1=\textbf{0}$ through JFM-trained U-Net (over $200$ steps) to obtain the noise $x_0$ that specifies the class-independent variation.  We then flow forward (over $200$ steps) using the same $x_0$ while changing $y_0$ to generate respective images of different classes. Inspecting what changes and what stays the same allows us to interpret the class-related and class-independent aspects of an image, as determined by the model. For example, in the second row, the hand is mostly left intact across images of different class, just as is the orange colour in the bottom row. The model clearly does not ``see" these aspects as class-related. On the other hand, the switch from uniform to horizon separated background in bottom row indicates that the ``ship" and ``truck" classification is strongly correlated with background. Consistency of generation and classification in JFM provides natural means of interpreting model's decision-making without post-hoc methods.

\begin{figure}[t!]
    \centering
    \includegraphics[width=0.95\textwidth]{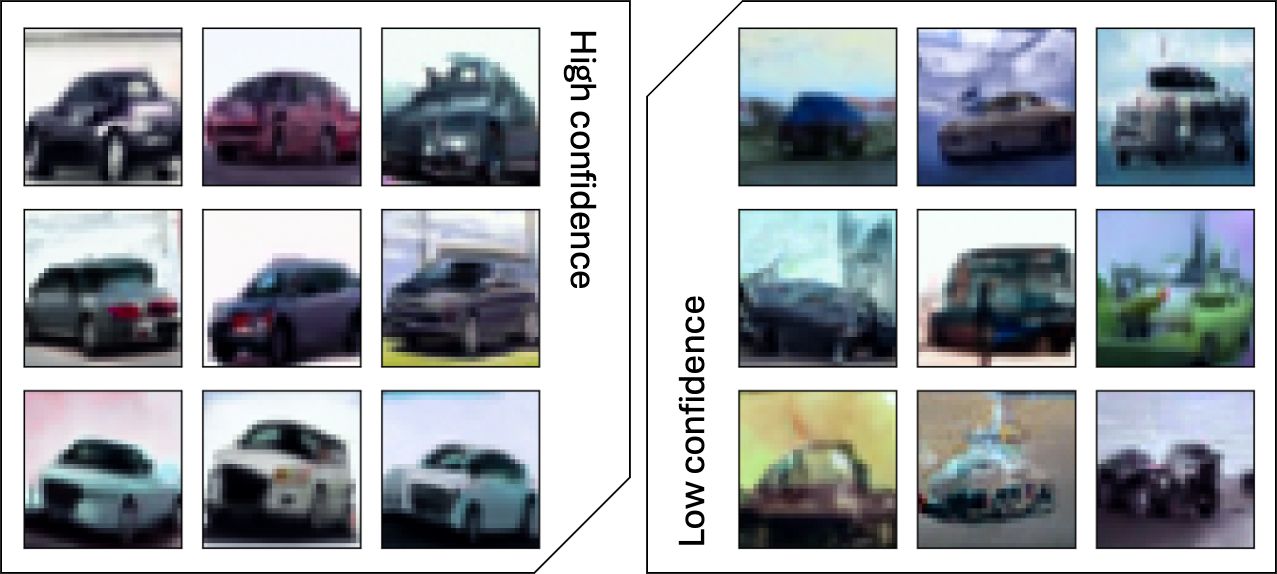}
    
    \caption{Sample of similar-looking images generated by a CIFAR-10-trained JFM conditioned on the "car" class, grouped by the confidence of the reverse-flow classification. Images on the left are classified as "car" with $\ge 0.9$ confidence, while images on the right are classified as "car" with $\le 0.1$ confidence. The consistency property allows classification confidence to be used to examine visual patterns associated with confident and uncertain generations of the flow model.}
    \label{figure: cifar10 generated confidence examples}
\end{figure}

Figure \ref{figure: cifar10 generated confidence examples} illustrates another potential interpretability avenue enabled by JFM. We generated images conditioned on a class label by sampling $x_0$ and forward flowing through JFM for 100 steps. Each generated image was then reverse flowed and classified 20 times, sampling a different $y_1$ on each pass to estimate the model's classification confidence. We binned the generated images by confidence and inspected them for systematic visual patterns associated with confident and uncertain predictions. In Figure \ref{figure: cifar10 generated confidence examples}, for example, images classified with high confidence as cars consistently exhibit a prominent front and rear light clusters, while lower-confidence examples tend to lack these features. JFM therefore provides a model-driven route to investigating learned behaviour: rather than first formulating a hypothesis about which features the model uses and then designing an experiment to test it, we can generate examples and inspect the patterns that emerge across confidence levels. The model itself brings potential edge cases and discriminative patterns to our attention. While JFM does not guarantee that its internal representation aligns with the semantic meaning of a class label, its generation and classification capabilities provide a direct means of probing the relationship between what the model generates and what it recognises. More examples are provided in Appendix \ref{appendix: cifar10 results}.

\section{Conclusion}

Joint Flow Matching (JFM) is a training framework for continuous normalising flows that yields consistent generative and discriminative inference from a single reversible model. By assigning $X$ and $Y$ to opposite temporal endpoints, the endpoint distributions factorise by construction, and the learned flow implicitly defines a joint distribution $p_\text{JFM}(X, Y)$. No additional constraints, regularisation, or auxiliary networks are required to enforce this consistency.

Consistency does not guarantee a better underlying generative model. Well-tuned hyperparameters, representative training data, and sufficient training are all still required to produce a useful model under JFM. However, JFM's consistency enables model-driven analysis of its learned behaviour, allowing the model to generate samples and assess its own confidence in them without requiring specific hypotheses about what to investigate.





This work has limitations. The slice-diffeomorphism assumption required for Equation \ref{equation: joint distribution} may be violated by pathological datasets. The framework also extends naturally to continuous $Y$ variables (e.g. regression targets or structured outputs), although the behaviour of the reversed conditional $p_\text{JFM}(Y | X)$ in such settings has not been characterised. Lastly, the training of a joint generative-discriminative model may be computationally more taxing than doing either separately.


Several extensions naturally follow from the JFM framework. The consistency guarantee holds for any choice of velocity model and noise distribution, so rectification \citep{LiuEtAl2022}, optimal-transport couplings \citep{TongEtAl2023}, and Riemannian generalisations \citep{ChenLipman2024} can all be applied directly. The joint distribution also offers a principled basis for model-internal interpretability: visualising $p_\text{JFM}(X | y_0)$ across different label values can expose which regions of the image distribution are associated with each class without post-hoc attribution methods. 

\vspace{-0.2cm}
\paragraph{Code availability.}
The implementation accompanying this work is publicly available at: \url{https://github.com/4c4a394f/anon-4c4a394f}.


\begin{ack}
This work was supported by a University of Otago University Research Grant (UORG).
\end{ack}



\bibliography{bibliography.bib}
\bibliographystyle{unsrtnat}

\appendix
\section{Reverse flow in the Spiral Example}
\label{appendix: reverse spiral} 

\begin{figure}[h!]
    \centering
    \begin{minipage}{0.45\textwidth}
        \begin{subfigure}[t]{\textwidth}
            \centering
            \includegraphics[width=\textwidth]{figures/spiral/data.png}
            \caption{The label and data distributions.}
            \label{subfigure: reverse spiral data}
        \end{subfigure}
        \begin{subfigure}[t]{\textwidth}
            \centering
            \includegraphics[width=\textwidth]{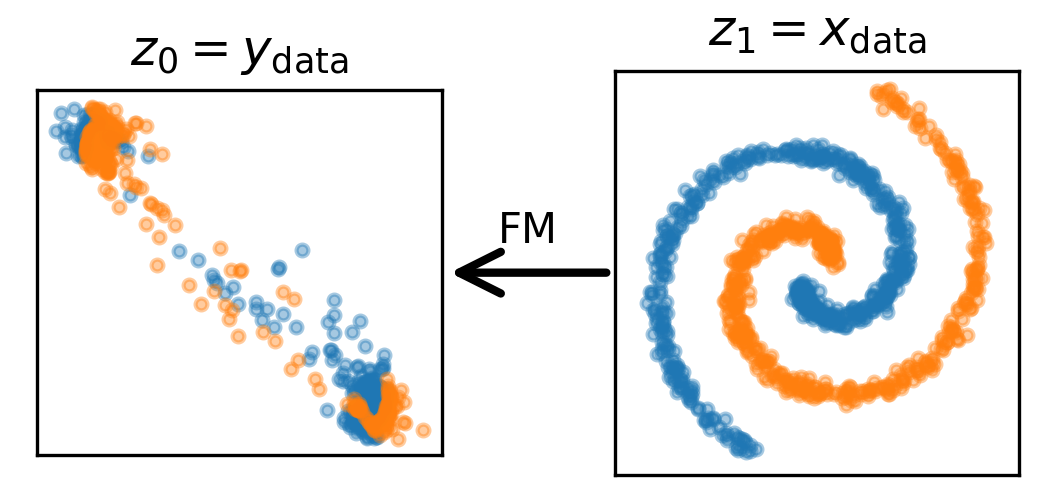}
            \caption{Standard flow matching classification.}
            \label{subfigure: reverse spiral fm}
        \end{subfigure}
    \end{minipage}%
    ~
    \begin{minipage}{0.5\textwidth}
        \begin{subfigure}[t]{\textwidth}
            \centering
            \includegraphics[width=\textwidth]{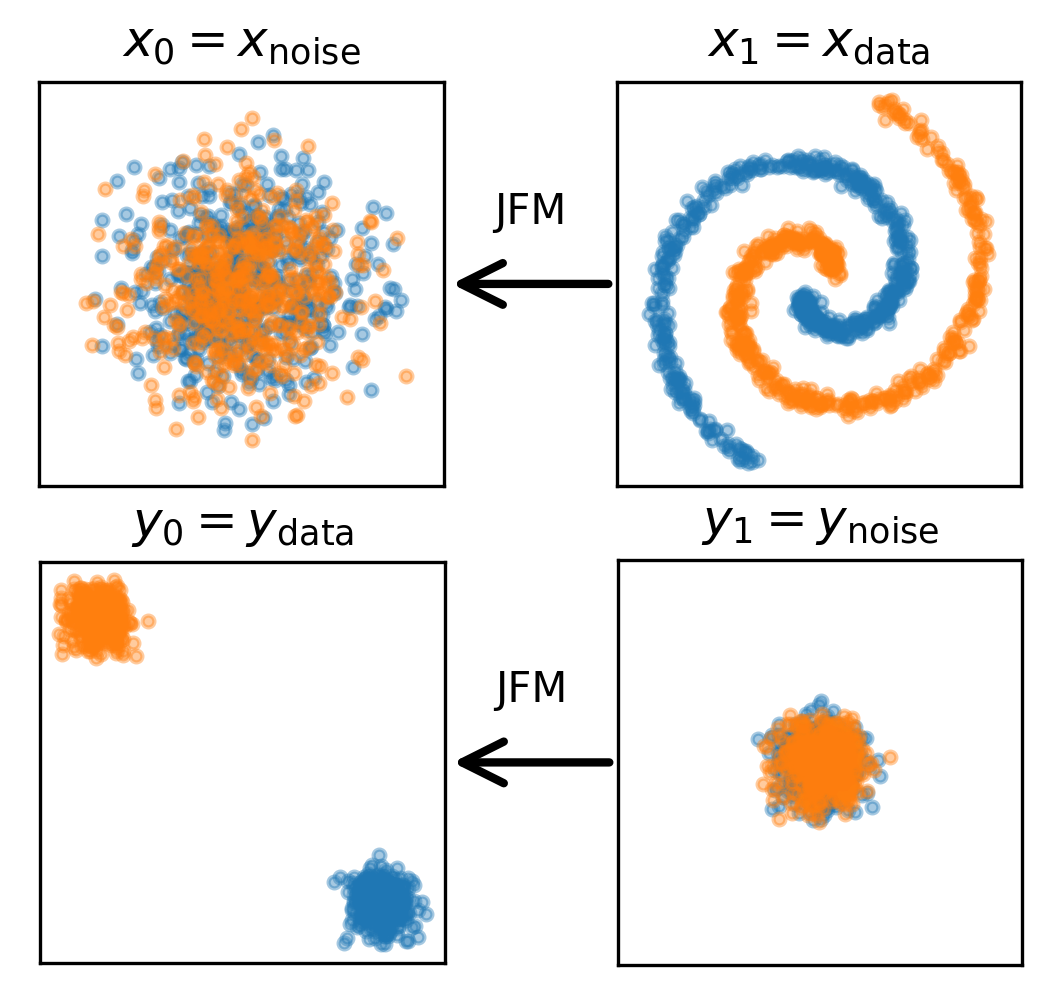}
            \caption{Joint flow matching classification.}
            \label{subfigure: reverse spiral jfm}
        \end{subfigure}
    \end{minipage}

    \caption{Reverse flow in FM and JFM trained to model $p(x|y)$ and $p(y|x)$ for the dataset shown in (\subref{subfigure: reverse spiral data}). Colour indicates the class conditioning, corresponding to the distinct arms of the spiral. In standard FM, consistent bidirectional conditioning requires $x$ to reverse flow directly to $y$; this works poorly for the spiral under linear interpolation path (\subref{subfigure: reverse spiral fm}). JFM has no trouble recovering the correct $p(y|x)$ under the same transport (\subref{subfigure: reverse spiral jfm}).}
    \label{figure: reverse spiral fm vs jfm}
\end{figure}

Figure \ref{figure: spiral fm vs jfm} showed the forward/generative flow, from labels $y_{\text{data}}$ to the spiral $x_\text{data}$ of the trained FM and JFM. Figure \ref{figure: reverse spiral fm vs jfm} shows the reverse/discriminative flow from the spiral to the labels. Classification would be determined by which cluster a given point ends up at.  

\section{CIFAR-10 U-Net architecture.}\label{appendix: unet}

\begin{figure*}[t]
\centering
\resizebox{\textwidth}{!}{%
\begin{tikzpicture}[
    >=stealth,
    node distance=0.35cm and 0.45cm,
    block/.style={
        draw,
        rounded corners=1pt,
        minimum height=0.9cm,
        align=center,
        font=\small
    },
    enc/.style={
        block,
        minimum width=1.55cm
    },
    dec/.style={
        block,
        minimum width=1.55cm
    },
    cond/.style={
        block,
        minimum width=2.0cm,
        minimum height=0.7cm
    },
    head/.style={
        block,
        minimum width=1.25cm,
        minimum height=0.65cm
    },
    arrow/.style={->,thick},
    skip/.style={
        ->,
        dashed,
        thick
    }
]


\node[enc] (enc1) {
    Encoder\\
    $32\times32$\\
    $128$
};

\node[enc, right=of enc1] (enc2) {
    Encoder\\
    $16\times16$\\
    $256$
};

\node[enc, right=of enc2] (enc3) {
    Encoder\\
    $8\times8$\\
    $256$
};

\node[enc, right=of enc3] (enc4) {
    Encoder\\
    $4\times4$\\
    $512$
};

\node[block, right=0.65cm of enc4, minimum width=1.8cm] (bottleneck) {
    Bottleneck\\
    $4\times4\times512$
};

\node[dec, right=0.65cm of bottleneck] (dec4) {
    Decoder\\
    $4\times4$\\
    $512$
};

\node[dec, right=of dec4] (dec3) {
    Decoder\\
    $8\times8$\\
    $256$
};

\node[dec, right=of dec3] (dec2) {
    Decoder\\
    $16\times16$\\
    $256$
};

\node[dec, right=of dec2] (dec1) {
    Decoder\\
    $32\times32$\\
    $128$
};

\node[block, right=of dec1, minimum width=1.25cm] (vx) {
    $v_x$
};


\draw[arrow] (enc1) -- (enc2);
\draw[arrow] (enc2) -- (enc3);
\draw[arrow] (enc3) -- (enc4);
\draw[arrow] (enc4) -- (bottleneck);

\draw[arrow] (bottleneck) -- (dec4);
\draw[arrow] (dec4) -- (dec3);
\draw[arrow] (dec3) -- (dec2);
\draw[arrow] (dec2) -- (dec1);
\draw[arrow] (dec1) -- (vx);


\node[block, left=0.7cm of enc1, minimum width=1.25cm] (xt) {
    $x_t$
};

\draw[arrow] (xt) -- (enc1);


\draw[skip]
    (enc1.north) -- ++(0,1.95)
    -| (dec1.north);

\draw[skip]
    (enc2.north) -- ++(0,1.55)
    -| (dec2.north);

\draw[skip]
    (enc3.north) -- ++(0,1.15)
    -| (dec3.north);

\draw[skip]
    (enc4.north) -- ++(0,0.75)
    -| (dec4.north);

\node[above=2cm of dec3] {
    skip connections
};


\node[cond, above=2.15cm of enc3] (yembed) {
    Label embedding\\
    $y_t:\;10\rightarrow512\rightarrow512$
};

\node[cond, above=of yembed] (tembed) {
    Time embedding\\
    $t:\;128\rightarrow512\rightarrow512$
};

\draw[arrow] ([xshift=-0.3cm]yembed.south) -- ([xshift=-0.3cm]enc3.north);
\draw[arrow] ([xshift=-0.3cm]tembed.south) -- ([xshift=-0.3cm]yembed.north);


\node[head, below=1.35cm of bottleneck] (flatten) {
    Flatten\\
    $8192$
};

\node[head, right=0.45cm of flatten] (fc1) {
    Linear\\
    $8192\rightarrow512$
};

\node[head, right=0.45cm of fc1] (fc2) {
    Linear\\
    $512\rightarrow256$
};

\node[head, right=0.45cm of fc2] (vy) {
    Linear\\
    $256\rightarrow10$
};

\node[block, right=0.45cm of vy, minimum width=1.25cm] (vy2) {
    $v_y$
};

\draw[arrow] (bottleneck.south) -- (flatten.north);
\draw[arrow] (flatten) -- (fc1);
\draw[arrow] (fc1) -- (fc2);
\draw[arrow] (fc2) -- (vy);
\draw[arrow] (vy) -- (vy2);

\node[above=0.15cm of vx] {
    image velocity
};

\node[below=0.15cm of vy2] {
    label velocity
};

\node[above=0.15cm of xt] {
    image
};

\node[left=0.15cm of yembed] {
    label
};

\node[above=0.15cm of bottleneck] {
    shared representation
};

\node[above=0.15cm of fc2] {
    classifier head
};



\end{tikzpicture}
}
\caption{
Architecture of the CIFAR-10 JFM model. The noisy image $x_t$ is processed by an ADM-style U-Net with four resolution levels. The continuous label vector $y_t$ and time $t$ are embedded and used to condition the U-Net. The bottleneck representation is used both by the decoder to predict the image velocity $v_x$ and by a separate MLP head to predict the label velocity $v_y$. The bottleneck has dimensions $4\times4\times512=8192$; the classification head maps this representation through dimensions $8192\rightarrow512\rightarrow256\rightarrow10$.
}
\label{fig:jfm_architecture}
\end{figure*}

The CIFAR-10 model uses an ADM-style U-Net with input dimensions $3\times32\times32$ and a base channel width of 128. The four resolution levels use channel multipliers $(1,2,2,4)$, giving channel widths of 128, 256, 256, and 512, respectively, with two residual blocks per level. Downsampling is performed between successive levels, giving a $4\times4$ bottleneck with 512 channels. Attention is applied at downsampling factors 1 and 2 using four attention heads. Residual blocks use GroupNorm with 32 groups, SiLU activations, scale-shift conditioning, and zero dropout.

The continuous label vector $y_t\in\mathbb{R}^{10}$ is embedded using a two-layer MLP with dimensions $10\rightarrow512\rightarrow512$ and combined with the corresponding time embedding. This conditioning is supplied throughout the U-Net. The bottleneck representation, of dimension $512\times4\times4=8192$, is flattened and passed to a classification head consisting of two hidden layers of 512 and 256 units, respectively. Each hidden layer uses LayerNorm and SiLU activation, with zero dropout, followed by a final linear layer producing the 10-dimensional label velocity $v_y$. The image velocity $v_x$ is produced by the U-Net decoder output. Thus, the label vector $y_t$ conditions the U-Net, while its velocity $v_y$ is predicted from the bottleneck representation.

The model was trained for 500 epochs with batch size 256.

\begin{figure}[t!]
    \centering
    \includegraphics[width=0.83\textwidth]{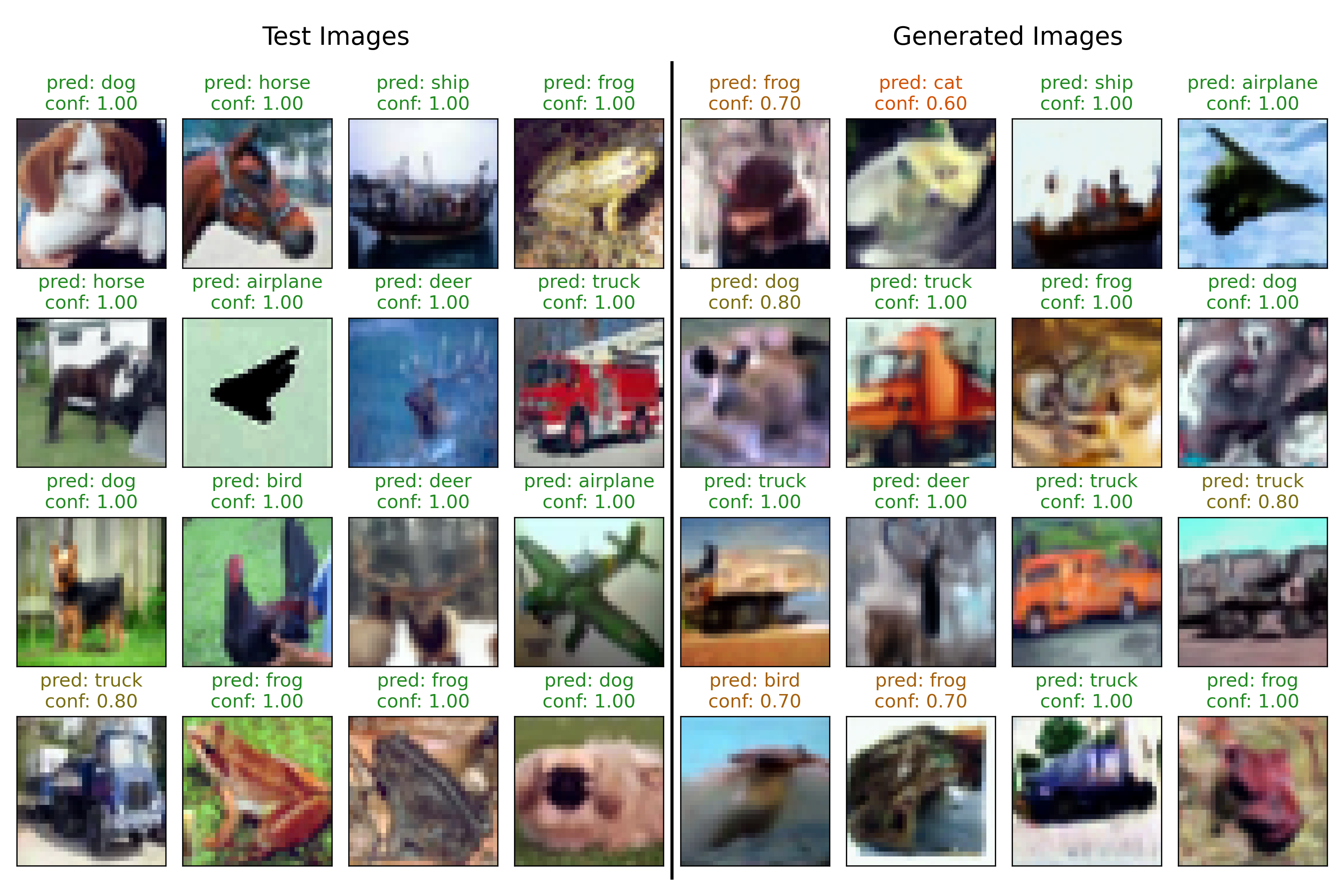}
    \caption{Classification decisions and confidence of the JFM trained on CIFAR-10 data evaluated on test (left side) and generated (right side) data. The colour of the label is colour-maped to green for high and red for low confidence.}
    \label{figure: cifar10 classification on test and generated examples}
    
\end{figure}

\section{CIFAR-10 Results}
\label{appendix: cifar10 results}

Figure \ref{figure: cifar10 classification on test and generated examples} provides a visualisation of generator-consistent classification on CIFAR-10. The left side shows classification decisions and confidence on a sample of test data by the model that achieved 0.908 test accuracy. The right-hand side shows classification decisions and confidence on a sample of data generated and classified via the forward and reverse flow of the JFM model.    

To assess sensitivity to the number of label-noise samples used to estimate classification confidence, we repeated the evaluation with 5, 10, and 100 samples. Accuracy was $0.906$, $0.907$, and $0.908$, with corresponding ECE values of $0.061$, $0.058$, and $0.055$, respectively. The small variation across sample counts indicates that the reported classification and calibration results are not strongly dependent on the number of samples used to estimate confidence.

\end{document}